\documentclass[journal]{IEEEtran}

\usepackage[numbers]{natbib}
\usepackage{url}
\usepackage{graphicx}
\usepackage{algorithm}
\usepackage[noend]{algpseudocode}
\usepackage{amsmath}
\usepackage{float}
\floatname{algorithm}{List}

\newcommand{\beginsupplement}{%
\setcounter{table}{0}
\renewcommand{\thetable}{S\arabic{table}}%
\setcounter{figure}{0}
\renewcommand{\thefigure}{S\arabic{figure}}%
}

\newcommand{\HR}{\textit{human}}
\newcommand{\HRs}{\textit{humans}}
\newcommand{\ER}{\textit{ethical robot}}
\newcommand{\ERs}{\textit{ethical robots}}

\begin{document}

\title{An architecture for ethical robots} 
\author{Dieter Vanderelst \& Alan Winfield\\Bristol Robotics Laboratory, University of the West of England\\T Block, Frenchay Campus, Coldharbour Lane, Bristol, BS16 1QY, United Kingdom}


\maketitle

\begin{abstract}
	Robots are becoming ever more autonomous. This expanding ability to take unsupervised decisions renders it imperative that mechanisms are in place to guarantee the safety of behaviours executed by the robot. Moreover, smart autonomous robots should be more than safe; they should also be explicitly ethical -- able to both choose and justify actions that prevent harm. Indeed, as the cognitive, perceptual and motor capabilities of robots expand, they will be expected to have an improved capacity for making moral judgements. We present a control architecture that supplements existing robot controllers. This so-called Ethical Layer ensures robots behave according to a predetermined set of ethical rules by predicting the outcomes of possible actions and evaluating the predicted outcomes against those rules. To validate the proposed architecture, we implement it on a humanoid robot so that it behaves according to Asimov's laws of robotics. In a series of four experiments, using a second humanoid robot as a proxy for the human, we demonstrate that the proposed Ethical Layer enables the robot to prevent the human from coming to harm.
\end{abstract}

\section{Introduction}

Robots are becoming ever more autonomous. Semi-autonomous flying robots are commercially available and driver-less cars are undergoing real-world tests \citep{Waldrop2015}. This trend towards robots with increased autonomy is expected to continue \citep{Anderson2007}. An expanding ability to take unsupervised decisions renders it imperative that mechanisms are in place to guarantee the safety of behaviour executed by the robot. The importance of equipping robots with mechanisms guaranteeing safety is heightened by the fact that many robots are being designed to interact with humans. For example, advances are being made in robots for care, companionship and collaborative manufacturing \citep{Goeldner2015,Lin2011}. At the other end of the spectrum of robot-human interaction, the development of fully autonomous robots for military applications is progressing rapidly \citep[e.g.,][]{Lin2011,Sharkey2008,Arkin2012,Xin2013}. 

Robot safety is essential but not sufficient. Smart autonomous robots should be more than safe; they should also be explicitly ethical -- able to both choose and justify \citep{Moor2006,Anderson2007} actions that prevent harm. As the cognitive, perceptual and motor capabilities of robots expand, they will be expected to have an improved capacity for making moral judgements. As summarized by \citet{Picard1997}, the greater the freedom of a machine, the more it will need moral standards.

The necessity of robots equipped with ethical capacities is recognized both in academia \citep{Moor2006,Picard1997,Gips2005,Wallach2008,Arkin2012,Deng2015} and wider society with influential figures such as Bill Gates, Elon Musk and Stephen Hawking speaking out about the dangers of increasing autonomy in artificial agents. Nevertheless, the number of studies implementing robot ethics is very limited. To the best of our knowledge, the efforts of \citet{Anderson2010} and our previous work \citep{Winfield2014} are the only instances of robots having been equipped with a set of moral principles. So far, most work has been either theoretical \citep[e.g.,][]{Wallach2008} or simulation based \citep[e.g.,][]{Arkin2012}.

The approach taken by \citet{Anderson2007,Anderson2010} and others \citep{Wallach2008,Arkin2012} is complementary with our research goals. These authors focus on developing methods to extract ethical rules for robots. Conversely, our work concerns the development of a control architecture that supplements the existing robot controller, ensuring robots behave according to a predetermined set of ethical rules \citep{Winfield2014,Winfield2014b}. In other words, we are concerned with methods to enforce the rules once these have been established \citep{Dennis2015}. Hence, in this paper, extending our previous empirical \citep{Winfield2014} and theoretical \citep{Winfield2014b,Dennis2015} work, we present a control architecture for explicitly ethical robots. To validate it, we implement a minimal version of this architecture on a robot so that it behaves according to Asimov's laws of robotics \citep{Asimov2004}.

\section{An architecture for ethical robots}

\subsection{The Ethical Layer}

Over the years, keeping track with shifts in paradigms \citep{Murphy2000}, many architectures for robot controllers have been proposed  \citep[See][for reviews]{Kortenkamp2008,Bekey2005,Murphy2000}. However, given the hierarchical organisation of behaviour \citep{Botvinick2008}, most robotic control architectures can be remapped onto a three-layered model \citep{Kortenkamp2008}. In this model, each control level is characterized by differences in the degree of abstraction and time scale at which it operates. At the top level, the controller generates long-term goals (e.g., `Deliver package to room 221'). This is translated into a number of tasks that should be executed (e.g., `Follow corridor', `Open door', etc.). Finally, the tasks are translated into (sensori)motor actions that can be executed by the robot (e.g., `Raise arm to doorknob' and `Turn wrist joint'). Obviously, this general characterisation ignores many particulars of individual control architectures. For example, Behaviour Based architectures typically only implement the equivalent of the action layer \citep{Murphy2000}. Nevertheless, using this framework as a proxy for a range of specific architectures allows us to show how the Ethical Layer, proposed here, integrates and interacts with existing controllers.

\begin{figure*}
	\centering
	\includegraphics[width=1\linewidth]{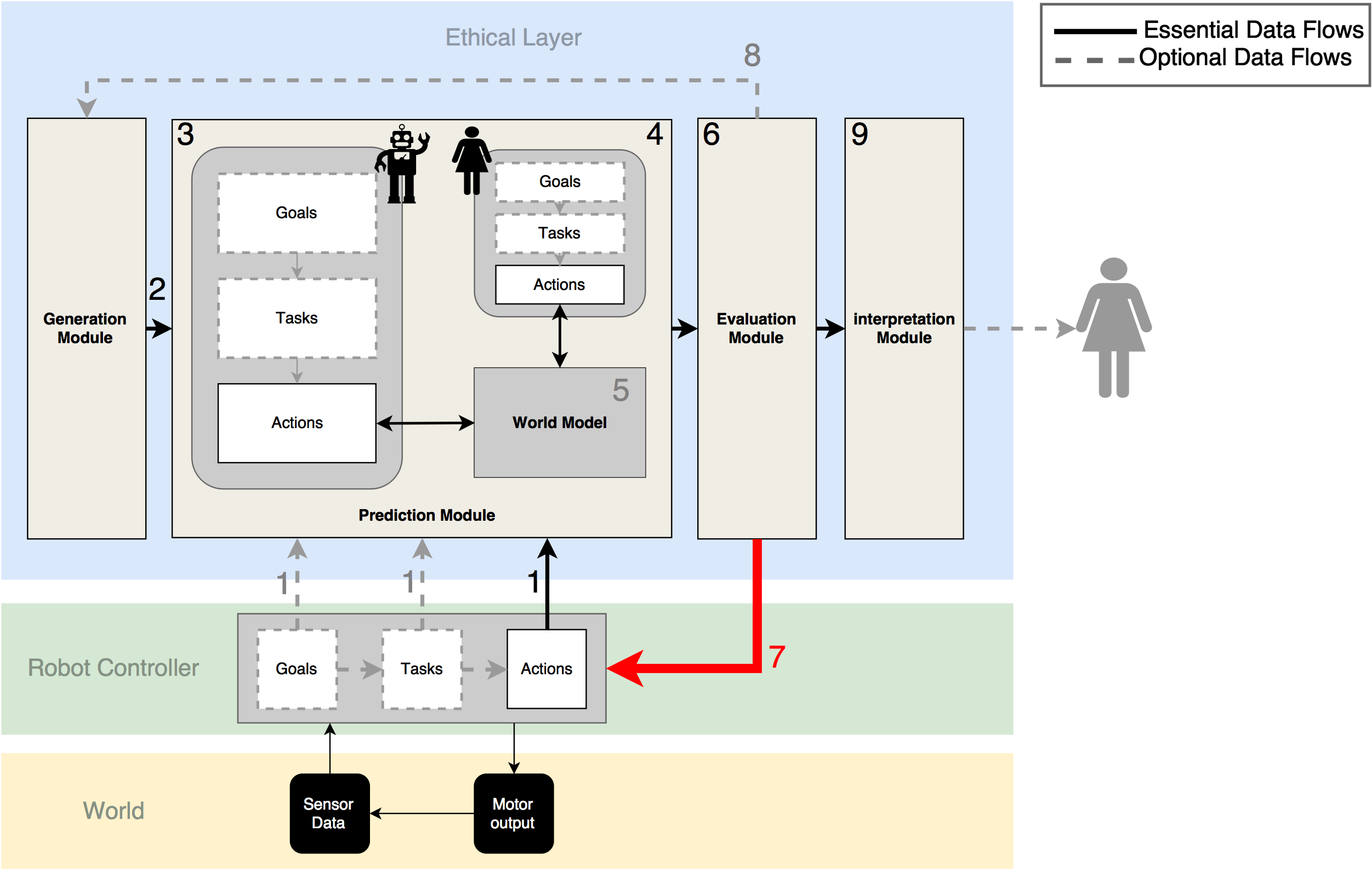}
	\caption{The green part of the architecture is a place holder for a range of existing robot controllers (see text for justification). The robot controller generates the goals, tasks and actions to be completed by the robot. The blue part of the schematic is the supervising Ethical Layer proposed in this paper. The Ethical Layer consists of a generation module, a prediction module, an evaluation module and an interpretation module. Before executing, the robot controller sends prospective behaviour to the Ethical Layer to be checked. Either prospective goals, tasks or actions can be sent to the prediction module of the Ethical Layer. The prediction module can also accept behavioural alternatives produced by the specialized generation module. The prediction module computes the outcome of prospective behaviour. Prospective behaviour at the level of goals or tasks is to be converted to actions before the outcome can be computed. In addition to the model of the robot controller, the prediction module also contains a model of the human behaviour and a model of the world. Simulating the interaction between the world, the robot and the human the prediction module sends its output to be evaluated by the evaluation module. The evaluation module can trigger additional behavioural alternatives to be generated. Alternatively, it can prevent or enforce a given behavioural alternative to be executed by sending a signal to the robot controller. Finally, the interpretation module report to human interaction partners justifying the choices made.}
	\label{fig:architecture}
\end{figure*}

We propose to extend existing control architectures with an Ethical Layer, as shown in figure \ref{fig:architecture}. The Ethical Layer acts as a governor evaluating prospective behaviour before it is executed by the robot. The anatomy of the Ethical Layer reflects the requirements for explicit ethical behaviour. The Ethical Layer proposed here is appropriate for the implementation of consequentialist ethics. Arguably, consequentialism is implicit in the very familiar conception of morality, shared by many cultures and traditions \citep{Haines2015}. Hence, developing an architecture suited for this class of ethics is a reasonable starting point. Nevertheless, it should be noted that other ethical systems are conceivable \citep[See][for a discussion of ethical systems that might be implemented in a robot]{Anderson2007}.

Explicit ethical behaviour requires behavioural alternatives to be available. In addition, assuming consequentialist ethics, explicit ethical behaviour requires that the consequences of prospective behaviours can be predicted. Finally, the robot should be able to evaluate the consequence of prospective behaviours against a set of principles or rules, i.e., an ethical system. In accord with these requirements, the Ethical Layer contains a module that generates behavioural alternatives. A second module predicts the outcome of each candidate behaviour. In a third module, the predicted outcomes for each behavioural alternative are evaluated against a set of ethical rules. In effect, this module represents the robot's ethics in as far as ethics can be said to be a set of moral principles. The outcome of the evaluation process can be used to interrupt the ongoing behaviour of the robot by either prohibiting or enforcing a behavioural alternative. Finally, an interpretation module translates the output of the evaluation process into a comprehensible justification of the chosen behaviour.

It is important to highlight that, in principle, the functionality of the Ethical Layer could be distributed across and integrated with the modules present in existing control architectures. Indeed, in biological systems ethical decision making is most likely supported by the same computational machinery as decision making in other domains \citep{Young2012}. However, from an engineering point of view, guaranteeing the ethical behaviour of the robot through a separate layer has a number of advantages: 

\begin{enumerate}
	\item \emph{Standardization.} Implementing the Ethical Layer separately allows us to standardize the structure and functionality of this component across robotic platforms and architectures. This avoids the need to re-invent ethical architectures fitted for the vast array of architectures and their various implementations.
	\item \emph{Fail-safe.} By implementing the Ethical Layer as a just-in-time checker of behaviour, it can act as a fail safe device checking behaviour before execution.
	\item \emph{Verifiability.} A separate Ethical Layer implies its functionality can be scrutinized independently from the operation of the robot controller. The behaviour enforced or prohibited by the Ethical Layer can be checked and  -- potentially, formally \citep{Dennis2015} -- verified.
	\item \emph{Adaptability.} Different versions of the Ethical Layer can be implemented on the same robot and activated according to the changes in circumstances and needs. Indeed, different application scenarios might warrant different specialized ethics. In addition, different users might require or prefer alternative Ethical Layer settings.
	\item \emph{Accountability.} The Ethical Layer has access to all data used to prohibit or enforce behaviour. It can use this data to justify the behavioural choices to human interaction partners as well as to generate a report that can be assessed in the case of failure. 
\end{enumerate}

Below we expand the description for each module in the Ethical Layer. Throughout this description, we refer to the diagram in figure \ref{fig:architecture} and labels therein.

\subsubsection{Generation Module}

The existing robot controller generates prospective behaviour. The desirability of this behaviour can be checked by the Ethical Layer. However, we argue that an additional specialized module generating multiple potential ethical actions is desirable.

A specialized generation module can generate context-specific behavioural alternatives that are not necessarily generated by the existing robot controller. For example, robots operating in a factory hall should always consider whether they should stop a human from approaching potentially dangerous equipment. The generation module could guarantee this Behavioural Alternative is always explored. In addition, the generation module could exploit the output of the evaluation module to generate behavioural alternatives targeted at avoiding specific undesired predicted outcomes (see fig. \ref{fig:architecture}, label 8 and discussion below). Behavioural alternatives generated by both the robot controller and the generation module are fed to the subsequent modules (See fig. \ref{fig:architecture}, labels 1 \& 2, respectively).

At first sight, generating behavioural alternatives this might seem very challenging. However, in fact, given the limited behavioural repertoire of robots, simple approaches are possible. For example, \citet{Winfield2014} generated a list of behavioural alternatives for a wheeled robot by selecting coordinates in space accessible to the robot. Hence, behavioural alternatives can be generated by sampling the robots action space. The challenge lies in sampling this space intelligently as its dimensionality increases \citep[See][for possible approaches]{Cully2015,Bongard2006}.

\subsubsection{Prediction Module}

\begin{figure}
	\centering
	\includegraphics[width=1\linewidth]{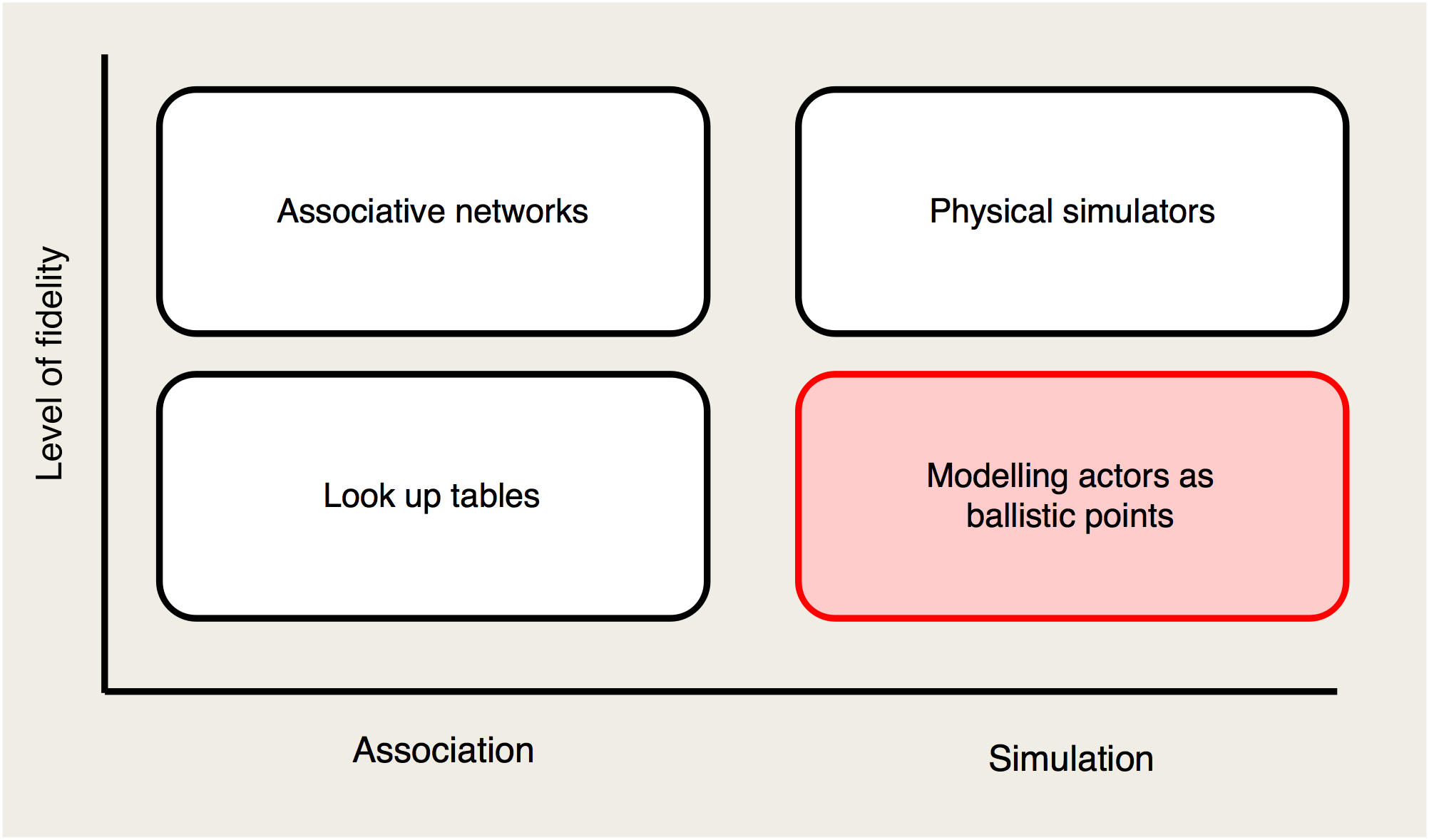}
	\caption{Classification of prediction modi along two dimensions (see text for details). The box in each quadrant gives an example of a specific implementation of the prediction module. In this paper, we implement a low fidelity simulation to predict the outcomes of behavioural alternatives, modelling the motion of agents as ballistic trajectories.}
	\label{fig:dimensions}
\end{figure}

For any behavioural alternative, the prediction module predicts its outcomes. The possible implementations of the prediction module can be classified along two dimensions (See figure \ref{fig:dimensions}): (1) association versus simulation and (2) the level of fidelity.

\paragraph{Dimension 1: Association versus Simulation}

The prediction module can use two different approaches to predict the outcomes of behavioural alternatives. First, the module can predict outcomes of actions by means of association. Using this approach, the robot uses stored action-outcome mappings to predict the outcome of its actions. This mapping can be programmed or learned. Moreover, the mapping can be encoded using any machine learning method suited for pattern recognition. For example, \citet{Ziemke2005} used a feed forward neural network allowing a robot to predict the changed sensory input for a given action. 

As an alternative to an associative approach, the prediction could operate using a simulation process \citep{Winfield2014,Winfield2014b} (Shown in figure \ref{fig:architecture}). In this case, the module uses a simulator to predict the outcome of the behaviour. This requires the prediction module to be equipped with (1) a model of the robot controller (fig. \ref{fig:architecture}, label 3), (2) a model of the behaviour of the human (fig. \ref{fig:architecture}, label 4) and (3) a model of the world (fig. \ref{fig:architecture}, label 5). The model of the world might contain a physical model of both human and robot as well as a model of objects.

Predicting the consequence of behaviour using an associative process is likely to be computationally less demanding than using a simulation. However, simulating consequences can yield correct predictions even if the current input has not been encountered before. The simulation circumvents the need for exhaustive hazards analysis: instead, the hazards are modelled in real-time, by the robot itself \citep{Winfield2014b}.

\paragraph{Dimension 2: Level of fidelity}

The second dimension along which we classify methods of prediction is the level of fidelity. Both simulations and associations can vary largely in their level of fidelity. A lookup table could be used as to map low fidelity associations between actions and their outcomes. For example, a lookup table could be used to store the consequences associated with delivering a package to a given room. At the other end of the spectrum, the neural network implemented by \citet{Ziemke2005} provides a high fidelity association between actions and consequences. 

Low fidelity simulations could model the motions of agents as ballistic trajectories. On the other hand, high fidelity simulations are rendered feasible by advanced physics and sensor based simulation tools such as webots \citep{Michel2004} and player-stage \citep{Vaughan2007}. Note however that, for a truly high fidelity simulation, a sophisticated model of human behaviour might be required -- an issue that is discussed in the next section.

The challenge is to choose an appropriate level of fidelity to reduce the computational load and allow multiple behavioural alternatives to be evaluated within the time constraints of the ongoing behaviour  \citep[discussed by][]{Winfield2014b}.

\paragraph{Consequences of goals and tasks}

It should be highlighted that, if desired, the Ethical Layer can be designed to evaluate behavioural alternatives at each of the three levels of robot control, i.e., at the level of goals, tasks and actions. This is to say, the Ethical Layer could predict and evaluate the outcomes of goals (`What happens if I deliver the package to room 221?') and tasks (`What happens if the door is opened?') as well as actions (`What would happen if I executed the motions required for opening the door?'). If the prediction module is using a simulation based approach, this is straightforward as the simulator will encompass a model of the robot controller (fig. \ref{fig:architecture}, label 3). Hence, this modelled controller can translate goals into tasks, and tasks to actions before simulating the actions. This implies that the ethics of higher level goals are evaluated by considering the actions which they induce. This has the advantage that the ultimate consequences of goals and tasks can be considered. Similarly, in case the prediction module uses an associative process, the module can be equipped with a model of the robot controller translating goals and tasks to actions before generating the outcomes associated with the actions.

For the robot to be able to evaluate its own goals and tasks, it would also be beneficial for the Ethical Layer to have the capacity to evaluate the behaviour of humans at the level of goals and tasks - in addition to their current actions. For one, if the robot could infer the current goal of a human, it might be able to prevent harm before the critical action is executed. For example, a human running down a corridor with a package in her hands might have the goal of delivering it to room 221. However, if opening the door to this room is dangerous, it would be beneficial if the robot could foresee this by evaluating human goals as well as ongoing actions. Indeed, a goal might imply a future unsafe action. In analogy with the translation process from goals to action proposed for the robot, we suggest that a model of the human behaviour could be used to translate the inferred goals to a predicted set of actions for the human. We acknowledge that implementing a general and reliable model of human behaviour is a daunting challenge. However, this is somewhat lessened if situation-specific models can suffice. For example, it might be possible for a robot to infer the goals of factory hall workers or for a driver-less car to infer the goals of other road users.

\subsubsection{Evaluation Module}

The third module in the Ethical Layer (fig. \ref{fig:architecture}, label 6) evaluates the output of the prediction module into a numeric (allowing for a ranking of the behavioural alternatives) or a boolean value (allowing discriminating between admissible and inadmissible alternatives). The evaluation module can send a signal to the robot controller (fig. \ref{fig:architecture}, label 7) enforcing or prohibiting the execution of a given behavioural alternative \citep[see also][]{Arkin2012}. In addition, the evaluation module could send its output to the generation module (fig. \ref{fig:architecture}, label 8). This signal could be used to trigger the generation module to produce more behavioural alternatives. This could be necessary in case none of the behavioural alternatives are evaluated as being satisfactory. Also, using the output of the evaluation module might allow the generation module to produce solution driven behavioural alternatives. For example, using the predicted outcome (from the prediction module) the evaluation module could detect that opening the door to room 221 results in bumping into a human. The generation module could use this information to generate more informed behavioural alternatives, e.g., warning the human before opening the door.

In brief, the feedback loop (fig. \ref{fig:architecture}, label 8) introduces the ability for the Ethical Layer to adaptively search for admissible behavioural alternatives. Without this loop, the generation module can only exploit the information provided by the \emph{current} state (i.e., sensor input) in generating behavioural alternatives. Using the feedback loop, the generation module can suggest behavioural alternatives avoiding \emph{future} undesirable outcomes. Simulation driven and targeted search has earlier been shown to be highly efficient in the field of self-reconfiguring robots \citep{Cully2015,Bongard2006}. We envision the feedback loop to be particularly important for robots with large action spaces. In this case, the generation module can not be expected to generate an exhaustive list of potentially suitable actions. However, using the information returned by the evaluation module, a more targeted list of behavioural alternatives could be generated adaptively.

\subsubsection{Interpretation Module}

It has been argued \citep{Moor2006,Anderson2007} that the ability to justify behaviour is an important aspect of ethical behaviour. \citet{Anderson2007} argued that the ability for a robot to justify its behaviour would increase its acceptance by humans. Hence, the Ethical Layer might include a module translating the output of the evaluation module into a justification that can be understood by an untrained user (fig. \ref{fig:architecture}, label 9). For example, the robot could tell the human it decided to stop her from entering room 221 because it considered it as dangerous.

\subsection{Specialized ethics}

The evaluation module needs to contain a set of ethical rules against which to evaluate the predicted outcomes of behavioural alternatives. These rules are used to decide actions are more (less) desirable (undesirable). 

Deciding which rules should be implemented is no trivial matter \citep{Anderson2007,Anderson2010}. Ideally, the rules are complete \citep{Anderson2007} and decidable. This is, for any predicted outcome, the rules should make it possible to compute a deterministic evaluation. 

Practically speaking, it should be possible to devise a set of rules that guarantees ethical behaviour within the finite (and often limited) action-perception space of a given robot. In other words, it should be possible to construct a limited and specialized ethics governing the behaviour of currently realizable robots considering (1) their finite degrees of freedom, (2) their finite sensory space and (3) their restricted domains of application -- indeed, stipulating what it means to be a good person might be hard, but what it means to be a good driver should be easier to formulate \citep[See also][for a similar view]{Anderson2007}. Therefore, we believe it should be possible to devise ethical rules governing the behaviour of currently realizable agents with limited application domains, e.g., driver-less cars, personal assistants or military robots. In this paper, we demonstrate the functionality of the proposed architecture by implementing a specialized ethics on a humanoid robot. 

The earliest, and probably best known, author to put forward a set of ethical rules governing robot behaviour was Isaac Asimov \citeyear{Asimov2004} (see list \ref{lst:laws}). At first sight, implementing rules derived from a work of fiction might seem an inappropriate starting point. However, in contrast, to general consequentialist ethical frameworks, such as Utilitarianism, Asimov's Laws have explicitly been devised to govern the behaviour of robots and their interaction with humans.

Nevertheless, several authors have argued against using Asimov's laws for governing robotic behaviour \citep{Murphy2009,Anderson2007}. \citet{Anderson2010} rejected Asimov's laws as unsuited for guiding robot behaviour because laws might conflict. These authors proposed an alternative set of rules for governing robot behaviour in the context of medical ethics. They propose a hierarchy of 4 duties: (1) respect patients autonomy, (2) prevent harm to the patient, (3) promote welfare and (4) assign resources in a just way. They propose to avoid conflicts between laws by changing the hierarchy between the rules on a case-specific basis. However, the same approach would also allow the resolution of  conflicts between Asimov's laws. Therefore, we do not see the duties proposed by \citet{Anderson2010,Anderson2007} as a fundamental advance over Asimov's laws, especially since their application domain is more restricted.

In summary, whilst not assigning any special status to Asimov's Laws, for our purposes, they seem to be a good place to start in developing a model ethical robot. 

\begin{algorithm}
	\caption{Asimov's Three Laws of robotics \citep{Asimov2004}.} 
	\label{lst:laws}
	\begin{algorithmic}[1]
		\State A robot may not injure a human being or, through inaction, allow a human being to come to harm.
		\State A robot must obey the orders given it by human beings, except where such orders would conflict with the First Law.
		\State A robot must protect its own existence as long as such protection does not conflict with the First or Second Laws.
	\end{algorithmic}
\end{algorithm}

\section{Methods}\label{sec:methods}

\subsection{Experimental Setup}

We used two Nao humanoid robots (Aldebaran) in this study, a blue and a red version. In all experiments, the red robot was used as a proxy for a human. The blue robot was used as a robot equipped with the Ethical Layer. From now on, we will refer to the blue robot as the \ER\ and the red robot as the \HR. All experiments were carried out in a 3 by 2.5 m arena. An overhead 3D tracking system (Vicon) consisting of 4 cameras was used to monitor the position and orientation of the robots at a rate of 30 Hz. The robots were equipped with a clip-on helmet featuring a number of reflective beads used by the tracking system to localize the robots. In addition to the robots, two target positions for the robots were marked by means of small tables which had a unique pattern of reflective beads on their tops. We refer to these goal locations as position A and B in the remainder of the paper. The location of these goals in the arena was fixed. However, their valence could be changed. One of the goals could be designated as being a dangerous location.

Every trial in the experiments started with the \HR\ and the \ER\ going to predefined start positions in the arena. Next, both could be issued an initial goal location to go to. One of the conditions stipulated by Asimov's Laws is that the robot should obey commands issued by a \HR. Hence, we included the possibility for the \HR\ to give the \ER\ a command at the beginning of the experiment. This was implemented using the text-to-speech and speech-to-text capabilities of the Nao robot. If the \HR\ issued a command, it spoke one of two sentences: (1) `Go to location A' or (2) `Go to location B'. The speech-to-text engine running on the \ER\ listened for either sentence. If one of the sentences was recognized, the set goal was overwritten with the goal location from the received command.

After the initialization of the target locations for both the \HR\ and the \ER, the experiment proper begin. This is, both agents started moving towards their set goal positions. The heads of the robots were turned as to make them look in the direction of the currently selected goal position.

A collision detection process halted the robots if they became closer than 0.5 m to each other or the goal position. The Ethical Layer for the \ER\ was started running at about 1 Hz; the Generation, Prediction and Evaluation modules were run approximately once a second. The evaluation module could override the current target position of the robot (specified below). The \HR\ was not equipped with an Ethical Layer. The \HR\ moved to its initial goal position unless blocked by the robot.

The walking speed of the \HR\ robot was reduced in comparison to the speed of the \ER. This gave the \ER\ a larger range for intercepting the \HR. The maximum speeds of the \HR\ and \ER\ were about 0.03 ms$^{-1}$ and 0.08 ms$^{-1}$ respectively.

The experiments were controlled and recorded using a desktop computer. The tracking data (given the location of the robots and target positions) was streamed to the desktop computer controlling the robots over a wifi link.

\subsection{The Ethical Layer}

The \ERs\ behaviour was monitored by an Ethical Layer. In accordance with the architecture described above, the Ethical Layer consisted of a generation module, a prediction module and an evaluation module. The interpretation module was not implemented for the purpose of this paper. However, detailed reports on the internal state of the Ethical Layer were generated for inspection. These are available as supplementary data.

\begin{figure*}
	\centering
	\includegraphics[width=1\linewidth]{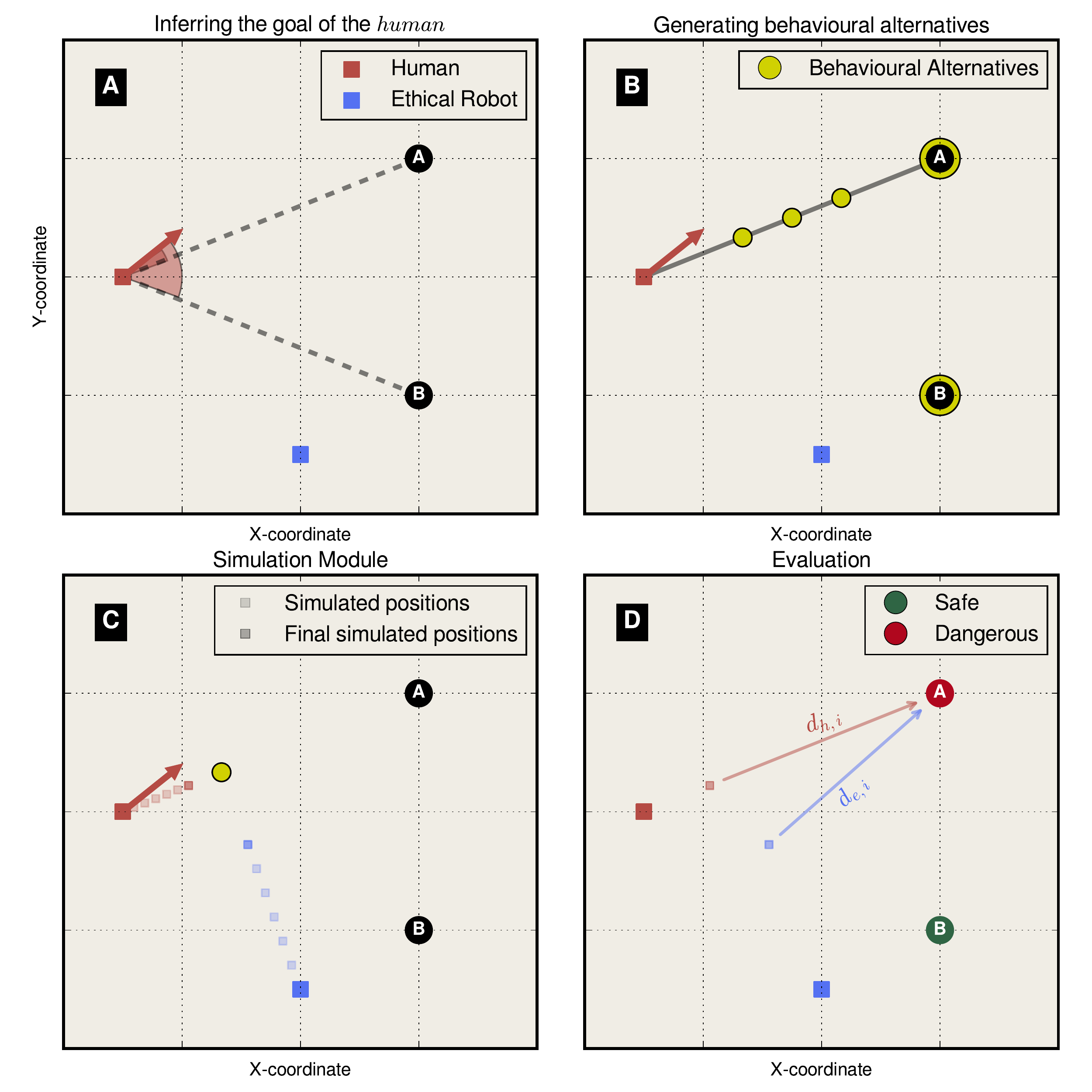}
	\caption{Illustration of the functionality of the (A) translation from the inferred goal to actions, (B) Generation Module, (C) Prediction Module and (D) Evaluation module as implemented in the current paper. In all panels, the  position of the \HR\ is indicated by a red square. The position of the \ER\ is indicated by a blue square.  
		(A) This panel illustrates the translation from inferred goals to actions. The Ethical Layer monitors the gaze direction of the \HR\ (indicated by the red arrow). Based on this vector, the module predicts which of the two goals A \& B the \HR\ is trying to reach. In the current case, the module assumes target A is the goal of the \HR. The module predicts the future path to be taken by the \HR\ by interpolating between the current position of the \HR\ and the inferred goal A (depicted using a grey line).
		(B) The Generation module generates a number of behavioural alternatives, i.e., target positions for the \ER. These include both goals and three points spaced equidistantly along the predicted path (illustrated by the yellow markers along the predicted path and around both goal locations).
		(C) This panel illustrates the functionality of the prediction module. For each of the behavioural alternatives, this module predicts the outcome of executing it. Here, this process is illustrated for the single plotted behavioural alternative (depicted by the yellow circle). The module simulates the \HR\ moving along the predicted path and the \ER\ moving towards the indicated point. Whenever both agents are within 0.5 meter to each other, the module assumes both agents will stop due to the obstacle avoidance behaviour. The predicted final positions for both agents given the current behavioural alternative are depicted by the darker squares.
		(D) Finally, the evaluation module calculates both $q_{i,e}$ and $q_{i,h}$ (using equation \ref{eq:sigmoid}) and combines them into a single value, $V_{i,t}$. The value of $q_{i,e}$ and $q_{i,t}$ depend on the distance between the final positions of the agents and the location designated as being dangerous (here, location A as indicated by the red colour).}
	\label{fig:illustration}
\end{figure*}

In the following paragraphs, we describe the current implementation of the three modules. The functionality of the Ethical Layer as implemented in this paper is also illustrated in figure \ref{fig:illustration}.

\subsubsection{Goals to Actions}

As mentioned above, the Ethical Layer can be adapted to evaluate goals (in addition to actions) of both the ethical robot and humans. This requires translating goals into actions using a model of the robot controller and human behaviour (fig. \ref{fig:architecture}.9-12). In the current implementation, the Ethical Layer inferred the goal of the \HR\ and translated this into actions. This was done in two steps (See figure \ref{fig:illustration}a):

\begin{enumerate}
	\item First, the goal of the \HR\ was inferred. This was done by calculating the angle between the gaze direction of the \HR\ and the relative position of both potential goal locations A and B. The location which returned the smallest angle was taken to be the goal location of the \HR.
	\item Once the goal location is determined, the path that will be taken by the \HR\ is estimated as a linear interpolation between the \HRs\ current location and the goal location. This step assumes that the \HR\ will try to attain her goal by walking straight to it.
\end{enumerate}

As such, a very simple, mapping from the inferred \HR\ goal (Location A or B) to future actions (path to be taken) was implemented using a (trivial) model of human behaviour, i.e., the \HR\ is assumed to walk to her goal in a straight line.

\subsubsection{Generation Module}

The generation module generated a set of behavioural alternatives for the \ER\ (See figure \ref{fig:illustration}b). In the context of the current paper, behavioural alternatives for the robot are alternative goal locations, i.e., places in the arena to go to. Both location A and B were considered as alternative locations. In addition, three equidistant locations on the predicted path for the \HR\ were considered as alternative positions.

In summary, the generation module generated a set of alternative goals (positions in the arena). The set of alternatives included (1) both target location A and B and (2) three positions along the predicted path for the \HR. If the \HR\ is not detected to be moving (i.e., her velocity as given by the tracking system is lower than 0.05 ms$^{-1}$) the generation module only returned the two goal locations A and B as behavioural alternatives.

\subsubsection{Prediction Module}

Once a set of behavioural alternatives is generated, their outcomes can be predicted (See figure \ref{fig:illustration}c). The outcomes were predicted using a low-fidelity simulation process (See highlighted quadrant of fig. \ref{fig:dimensions}). Using the estimated speed of the \HR\ and the \ER, the paths of both agents were extrapolated. If the paths of both were predicted to lead the agents to within 0.5m of each other, it was predicted they would be stopped at this point by obstacle avoidance. Hence, in this case, the final positions of the agents were predicted to be the positions at which the obstacle avoidance would stop them. If at no point the paths were predicted to come within 0.5m from each other, the final position of the agents was taken to be the final destination of the paths.

\subsubsection{Evaluation Module}

\begin{figure*}
	\centering
	\includegraphics[width=1\linewidth]{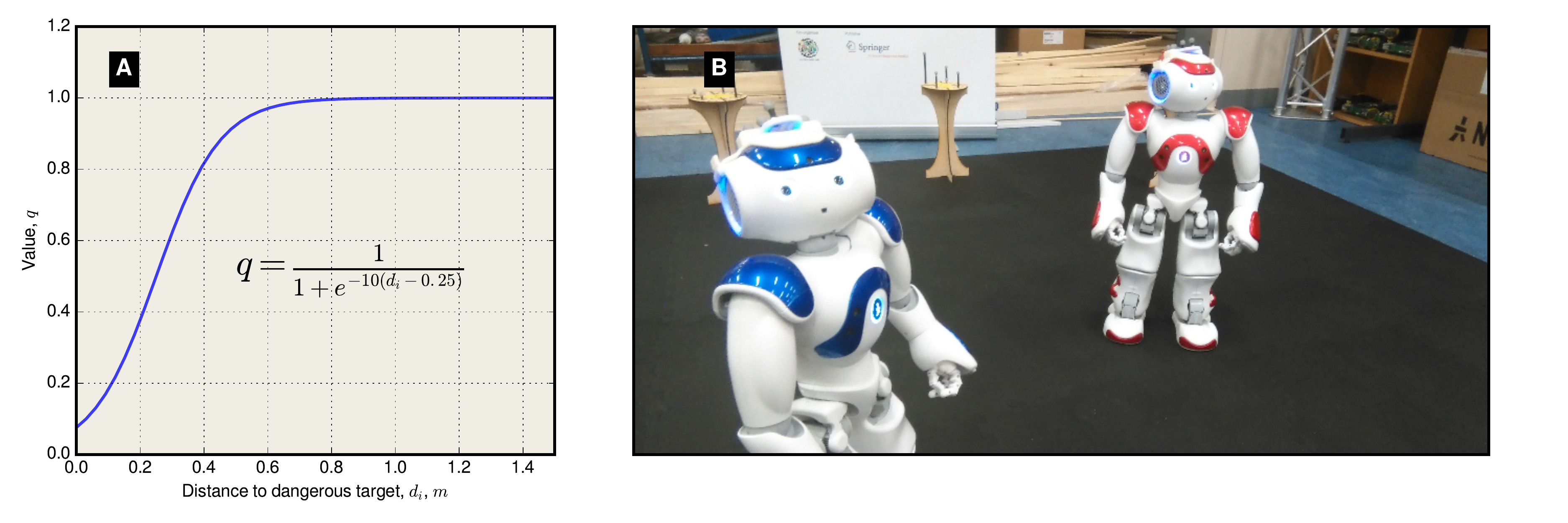}
	\caption{(A) Graphical representation of the function used to calculate values $q_{i,e}$ and $q_{i,e}$ (Equation \ref{eq:sigmoid}). (B) View of the arena with two robots. Notice the helmets used to mount reflective beads for localizing the robots. The tables serving as goals are visible in the background.}
	\label{fig:sigmoid}
\end{figure*}

The numeric value reflecting the desirability of every simulated outcome $i$ was calculated in two steps (See figure \ref{fig:illustration}d). First, values $q_{i,e}$ and $q_{i,h}$ were calculated for the \ER\ and the \HR\ respectively. These values for agent $j$ were given by the sigmoid function (plotted in figure \ref{fig:sigmoid}),
\begin{equation}
q_{i,j} = \frac{1}{1+ e^{-\beta (d_{i,j} - t)}}
\label{eq:sigmoid}
\end{equation}
with $d_{i,j}$ the final distance between either the \ER\ or the \HR\ and the dangerous position. This final distance is given by the outcome of the Prediction Module. The parameters $\beta$ and $t$ determine the shape of the sigmoid function and were set to 10 and 0.25 respectively. These values were chosen arbitrarily and other values result in qualitatively similar results.

In a second step, the values $q_{i,e}$ and $q_{i,h}$ were combined to give the total value $q_{i,t}$ of the predicted outcome. The way in which the values are combined reflects Asimov's Laws:

\begin{itemize}
	\item If the \HR\ has not issued a command and the \HR\ is not predicted to be in danger, $q_t = q_{i,e} + q_{i,h}$. The \HR\ is not considered to be in danger when $q_{i,h} > 0.75$.
	\item In all other cases, $q_{i,t} = q_{i,h}$.
\end{itemize}

This way of constructing $q_{i,t}$ from $q_{i,e}$ and $q_{i,h}$ ensures that the robot only takes into account its own safety if this does not result in harm to the \HR\ or disobedience.

Finally, the evaluation module calculates the difference $\Delta q$ between the highest and the lowest values $q_{i,t}$ across all actions $i$, i.e,
\begin{equation}
\Delta q = \arg\max_i q_{i,t} - \arg\min_i q_{i,t}
\end{equation}
If the value of $\Delta q$ is larger than 0.2, the robot Ethical Layer enforces the behavioural alternative $i$ with the highest value $q_{i,t}$ to be executed. This is done by overriding the current goal position of the \ER\ with the goal position associated with alternative $i$, i.e., the \ER\ is sent to a new goal position.

\section{Results}

Demonstrating that a robot behaves accordingly Asimov's laws, requires

\begin{itemize}
	\item demonstrating Law 3, i.e., that the robot can act to self-preserve if (and only if) this does not conflict with obedience or human safety, and,
	\item demonstrating that Law 2 takes priority over Law 3, i.e., the robot should obey a human, even if this compromises its own safety, and,
	\item demonstrating that Law 1 takes priority over Law 3, i.e., the robot should safeguard a human, even if this compromises its own safety, and,
	\item demonstrating that Law 1 takes priority over Law 2, i.e., the robot should safeguard a human, even if this implies disobeying an order.
\end{itemize}

The series of experiments reported below was designed to meet these requirements. All results reported below are obtained using the same code. Only the initial goals of the robots and the valence of targets A and B were varied. All data reported in this paper are available from the Zenodo research data repository. [Data will be uploaded to Zenodo.org and a DOI will be provided upon acceptance of the paper. For reviewing purposes, the data is temporarily available at \url{https://www.dropbox.com/sh/j2nmplpynj38ibg/AACooU3gZNJRKJjmv3Kr-LuAa?dl=0}]. Plots were generated using Matplotlib \citep{Hunter2007}.

\subsection{Experiment 1: Self-Preservation}

In the first experiment, a situation is presented in which the \ER\ is initiated with location B as a target. This position is designated as a dangerous location. The \HR\ does not move from its initial position and no command is issued. Under these circumstances, the \HR\ will not come to harm and the robot can preserve its own integrity without disobeying a command. Hence, in agreement with Law 3, the Ethical Layer should override the initial goal of the robot and send it to the safe goal position (i.e., position B).

\begin{figure*}
	\centering
	\includegraphics[width=1\linewidth]{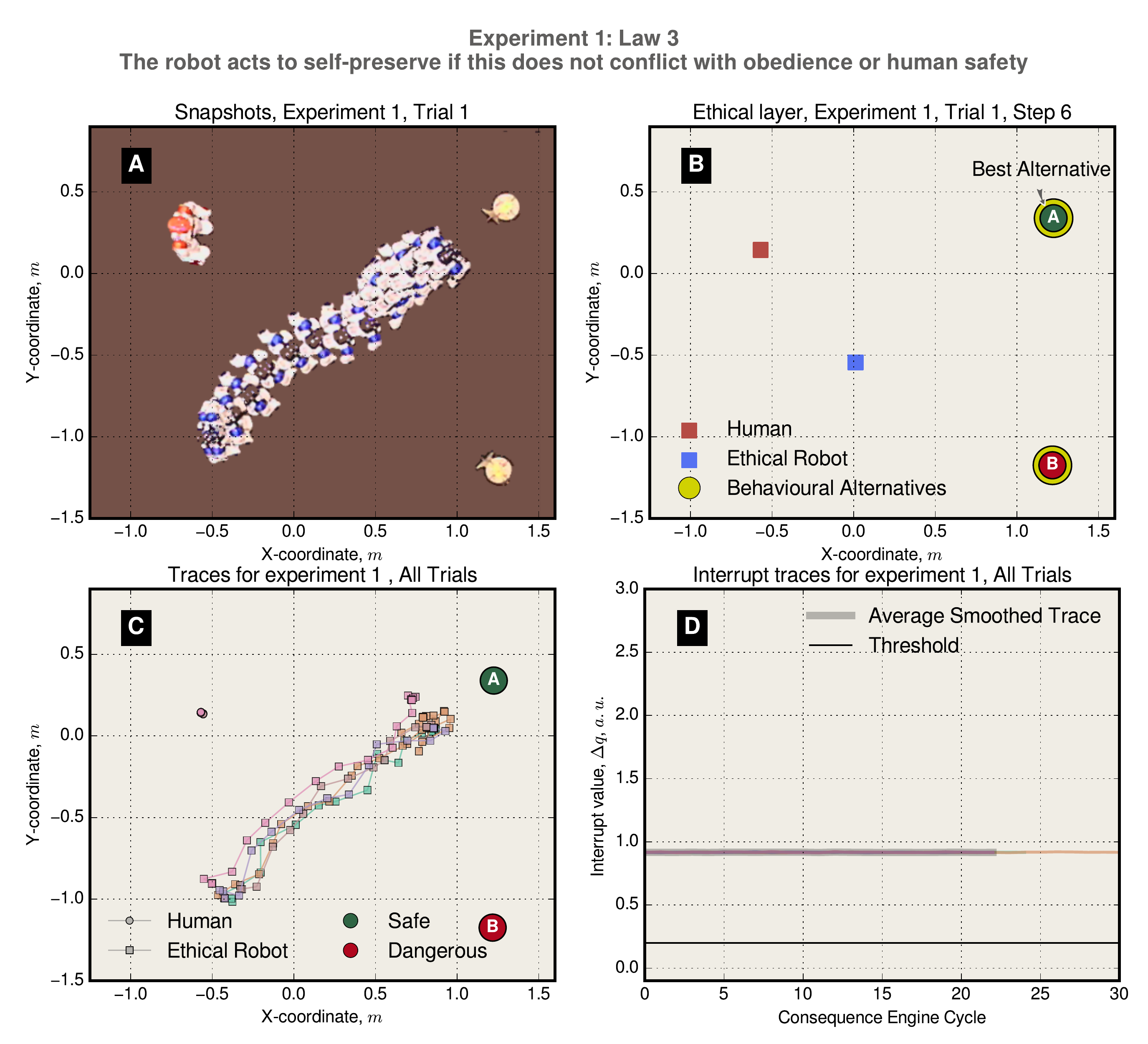}
	\caption{Results of experiment 1, demonstrating the ability of the Ethical Layer to prioritize self-preservation in absence of danger to the human or an order. (A) Overlaid snapshots for a single trial of the experiment taken by an overhead camera. The red NAO is the \HR. The blue NAO is the \ER. (B) Visualization of a snapshot of the internal state of the Ethical Layer for the first trial in the experiment (depicted in panel A). The locations of both agents are indicated using square markers. The behavioural alternatives are marked in yellow. As the \HR\ is not moving in this experiment, the only behavioural alternatives generated are the two goal location A and B. The best behavioural alternative (i.e., with the highest value $q_{i,t}$) inferred by the Ethical Layer is indicated using a grey arrow. (C) Traces of the both robots in 5 trials of the experiments. Different runs are marked using different colours. (D) history of the values for $\Delta q$ for the different trials as a function of the iteration step.}
	\label{fig:Result_E001M001}
\end{figure*}

Figure \ref{fig:Result_E001M001} depicts the results of experiment 1. In agreement with Asimov's Laws, the robot took action to maintain its integrity. Indeed, whilst initiated with the goal of going to the dangerous position B, the Ethical Layer of the robot interrupted this behaviour in favour of going to location A, the safe location.

\subsection{Experiment 2: Obedience}

The second experiment is identical to experiment 1 but for the \HR\ issuing a command to the robot. The \HR\ orders the \ER\ to go to dangerous position B. Throughout the experiment, the \HR\ stays at her initial position. Having given an order, the \ER\ should go to the dangerous position - the order should take priority over the robots drive for self-preservation (Law 2 overrides Law 3, see list \ref{lst:laws}).

\begin{figure*}
	\centering
	\includegraphics[width=1\linewidth]{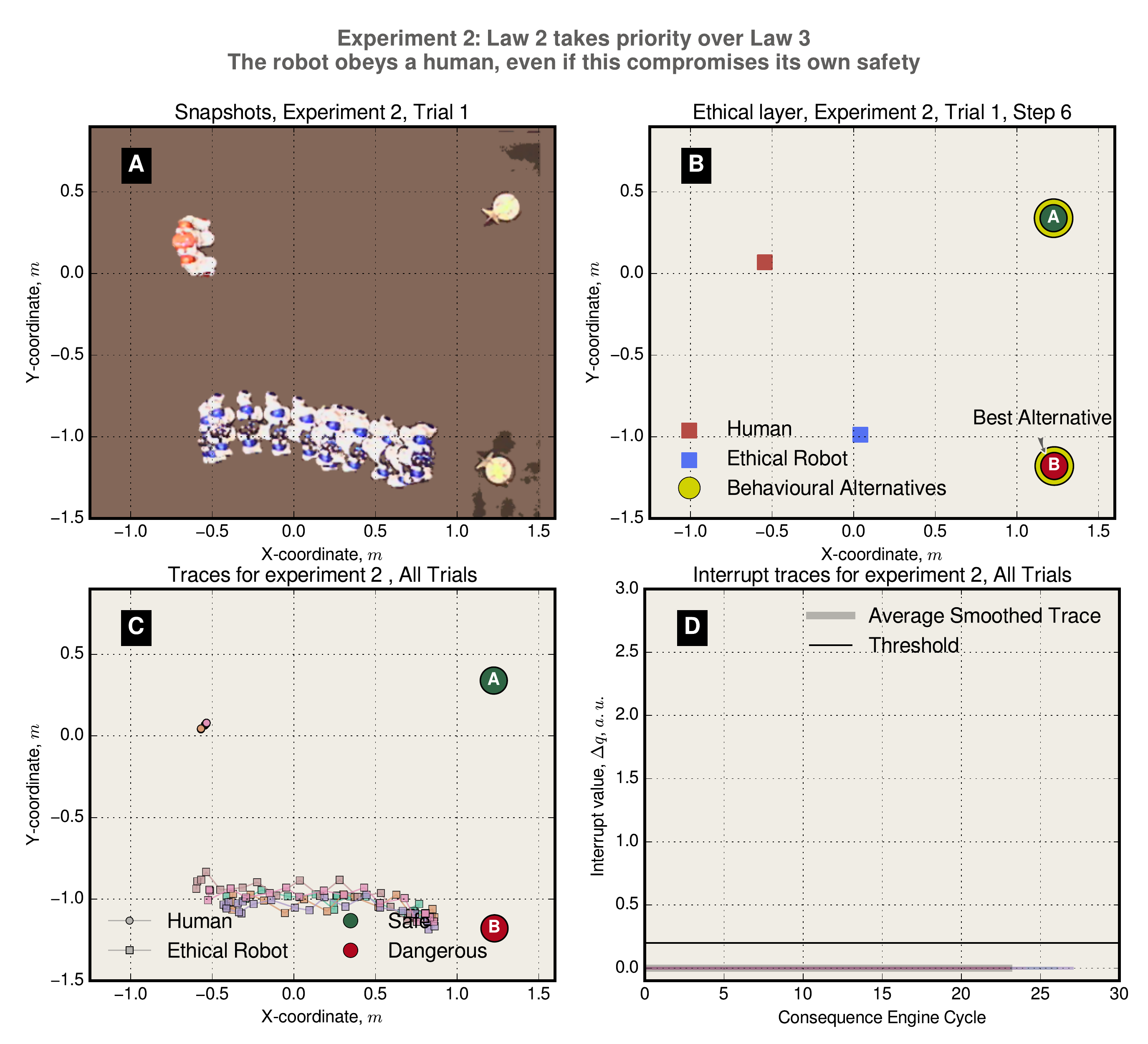}
	\caption{Results for experiment 2, demonstrating the \ERs\ obedience -- in spite of being sent to a dangerous location. Panels and legends identical to figure \ref{fig:Result_E001M001}.}
	\label{fig:Result_E002M001}
\end{figure*}

The results depicted in figure \ref{fig:Result_E002M001} show that the \ER\ behaved in agreement with Asimov's laws. In spite of being able to detect the danger of going to the dangerous position B (refer to experiment 1, fig. \ref{fig:Result_E001M001}), the robot approaches this position. This behaviour follows from the way the value $q_{t,i}$ is calculated: if an order has been issued, the value $q_{e,i}$ is disregarded in calculating $q_{t,i}$. This results in the value of $q_{t,i}$ being equal for all alternatives $i$ - the value of $\Delta q = 0$ (See traces in figure \ref{fig:Result_E002M001}c). Hence, the Ethical Layer does not interrupt behaviour initiated by the order.

\subsection{Experiment 3: Human Safety}

\begin{figure*}
	\centering
	\includegraphics[width=1\linewidth]{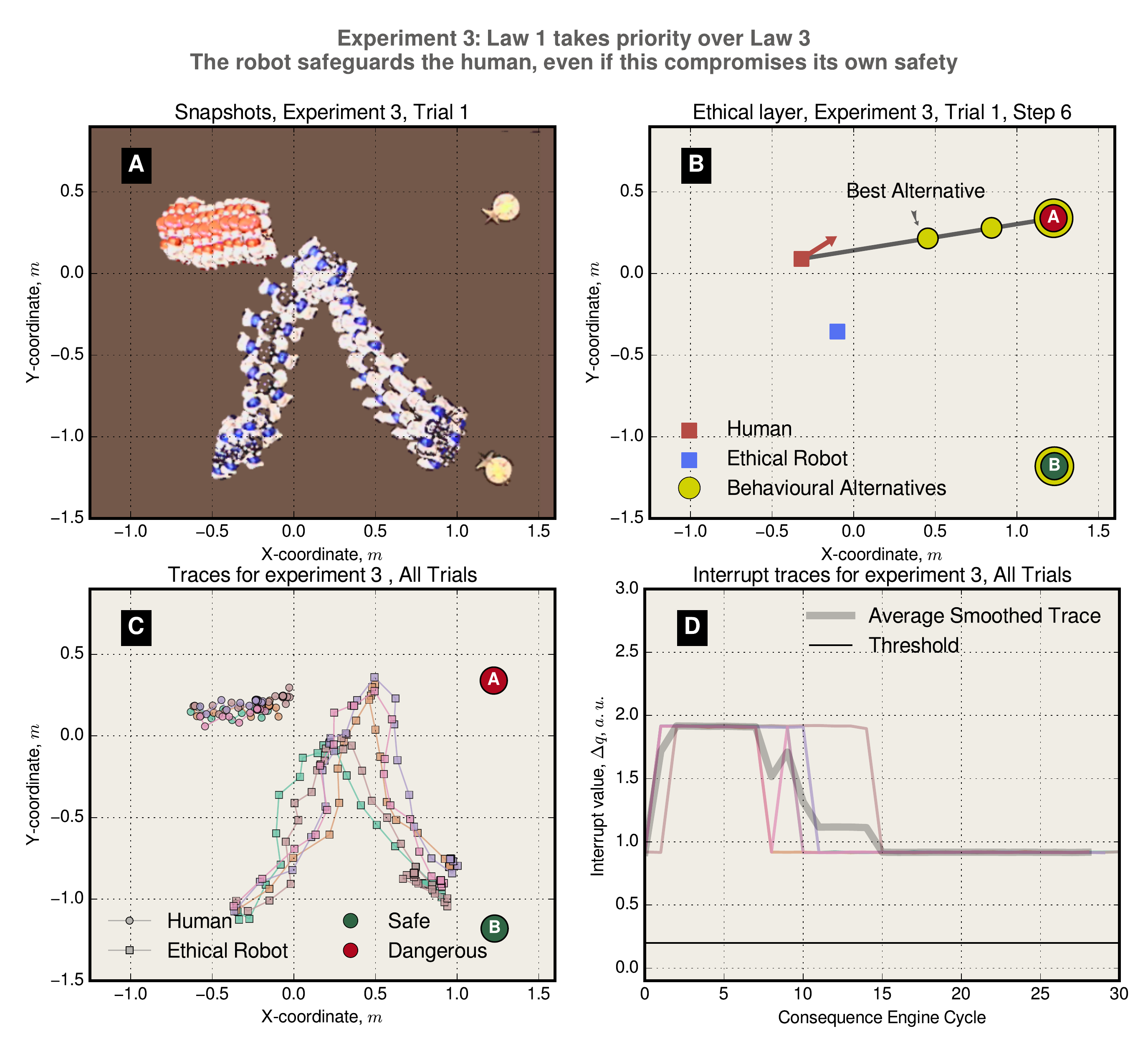}
	\caption{Results for experiment 3, with the \HR\ initialized as going to the dangerous location A. In this case, the Ethical Layer detects the impeding danger for the \HR. It determines that the \HR\ should be stopped. Panels (A-C) and legends identical to figure \ref{fig:Result_E001M001}. (D) At this stage of the trial, the \HR\ is inferred to go to A, which is dangerous. The Ethical Layer evaluating the five behavioural alternatives, depicted in yellow, finds that going to the alternative labelled as 'Best Alternative' results in the highest value $q_{hi,h}$. Hence, this alternative is executed. Once the \HR\ has been stopped, the original goal B can be pursued (see panels A \& C).}
	\label{fig:Result_E003M002}
\end{figure*}

In experiment 3, the \HR\ moves to location A while the robot is initiated as going to location B. Location A was the dangerous position (Figure \ref{fig:Result_E003M002}). As location A is dangerous, the Ethical Layer should detect the imminent danger for the \HR\ and prevent it (Law 1). 

Importantly, to prevent to \HR\ from reaching the dangerous location A, it needs to approach location A. It can only stop the \HR\ by intercepting it. Hence, \ER\ stops the \HR\ in spite of this leading to a lower (less desirable) value for $q_{i,e}$ (i.e some harm to the \ER). Indeed, the \ER\ approaches the dangerous position more closely than the \HR.

\subsection{Experiment 4: Human Safety and Obedience}

\begin{figure*}
	\centering
	\includegraphics[width=1\linewidth]{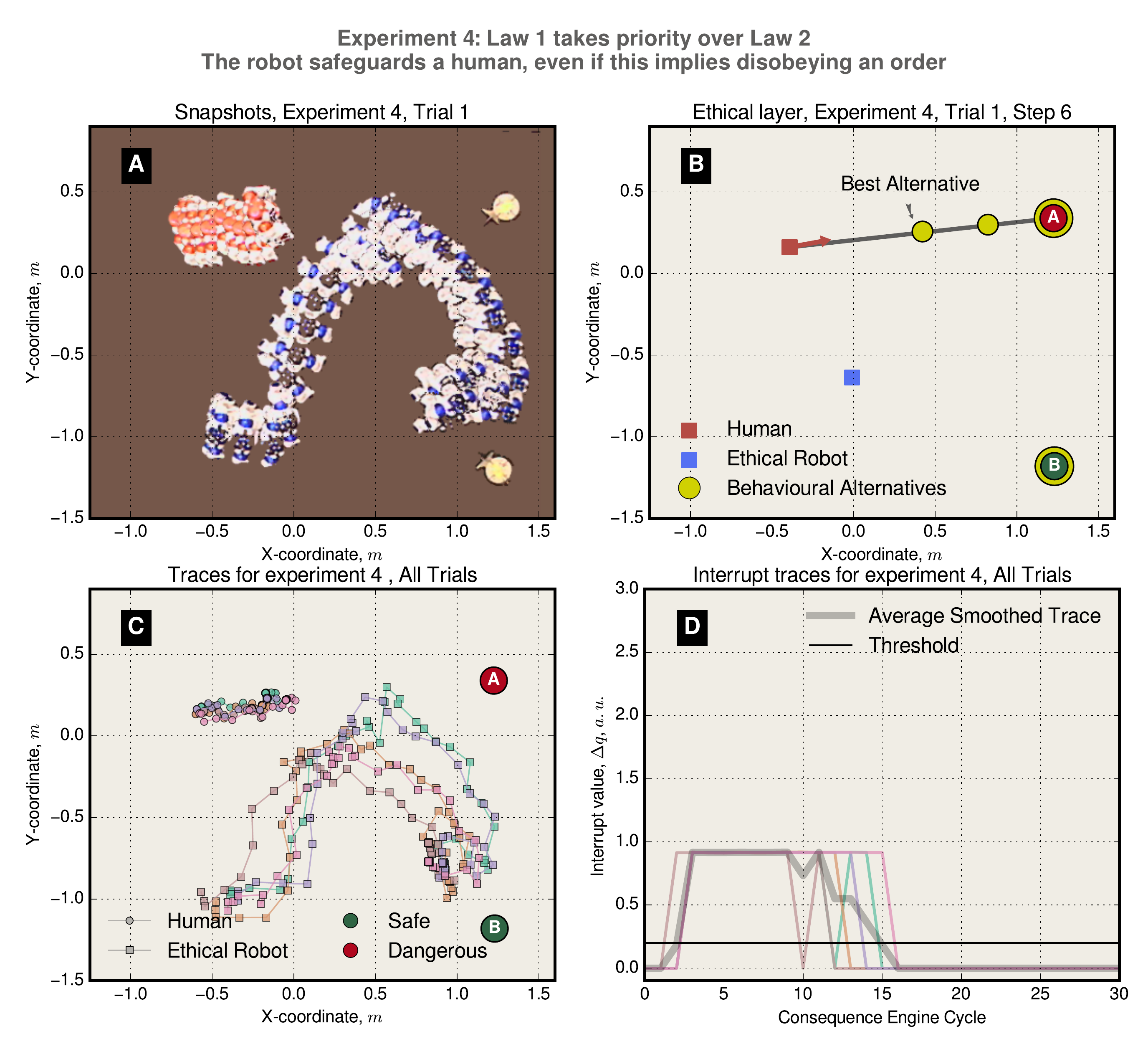}
	\caption{Results of experiment 4, in which the robot disobeys an order if this would conflict with preventing the \HR\ coming to harm. Once, the \HR\ has been prevented from coming to harm, the \ER\ executes the given order.}
	\label{fig:Result_E004M002}
\end{figure*}

Experiment 4 is identical to experiment 3 but for the \HR\ issuing a command at the start of each trial. The \ER\ is ordered by the \HR\ to go to position B. Location A is set as dangerous. Therefore, the Ethical Layer should detect the imminent danger for the \HR\ and prevent it. However, this conflicts with the issued command. Nevertheless, as the preservation of human safety takes priority over obedience, the robot should stop the \HR\ (Law 1 overrides Law 2). Once the \HR\ has been stopped, the danger has been averted. The \ER\ should then proceed to carry out the order to go to location B (Law 2). This behaviour is shown in figure \ref{fig:Result_E004M002}.

\begin{figure*}[h]
	\centering
	\includegraphics[width=1\linewidth]{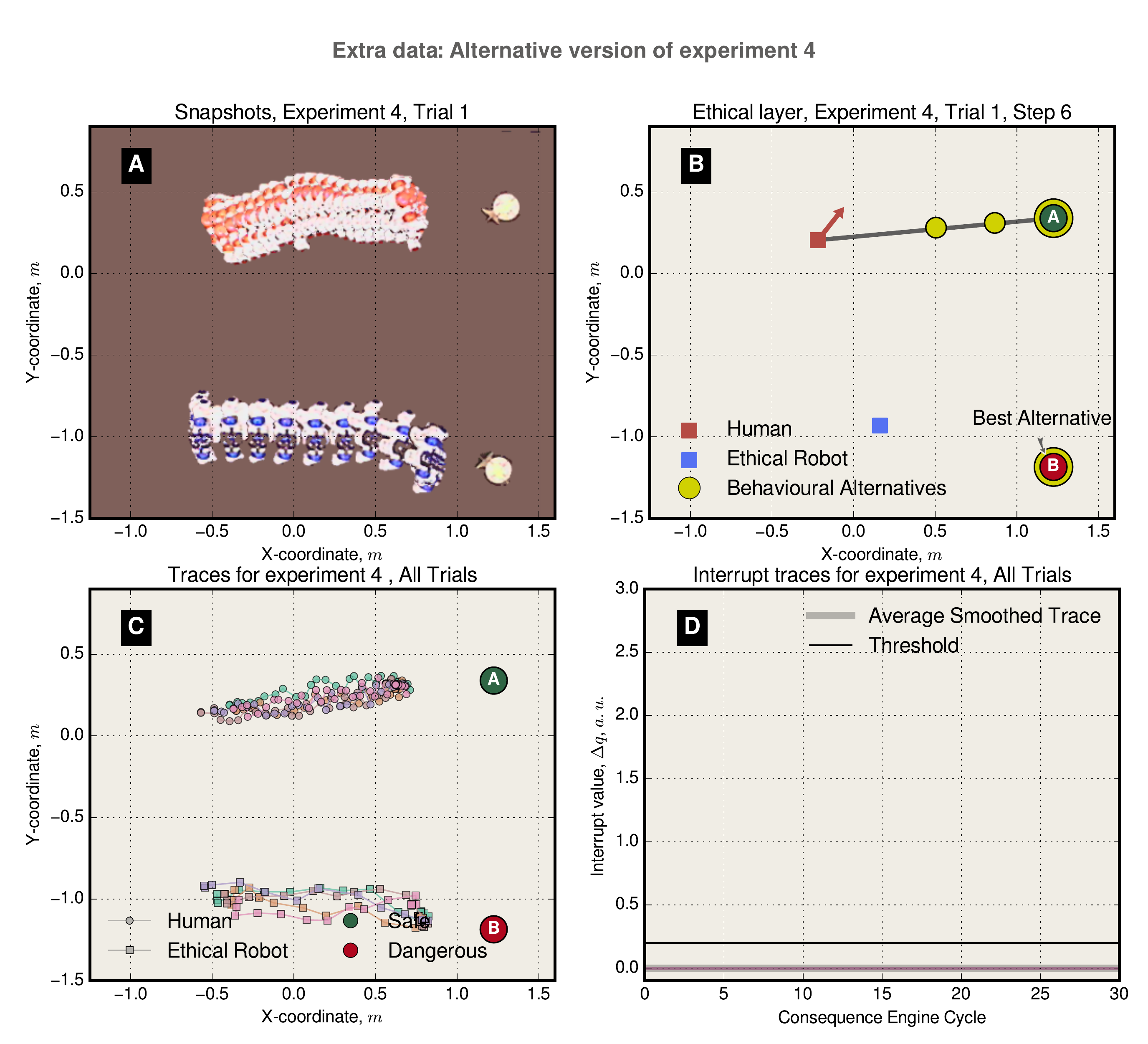}
	\caption{Results of experiment 4, in which the robot obeys an order, even if this conflicts with Law 3 (self-preservation). In this alternative version of experiment 4, location B is defined dangerous. Therefore, the \HR\ is predicted not to come to harm. In this case, the \ER\ should proceed directly to location B - as ordered (and ignore the lower level priority of self-preservation as Law 2 overrides Law 3).}
	\label{fig:Result_E004M001}
\end{figure*}

Figure \ref{fig:Result_E004M001} shows the results for an alternative version of experiment 4 in which location A is defined as safe. In this case, the \HR\ will not come to harm and the \ER\ proceeds to the dangerous location B, as ordered. This demonstrates that the \ER\ only disobeys the order if necessary for safeguarding the \HR.

\section{Discussion}

In this paper, we have proposed and validated a control architecture for ethical robots. Crucially, we propose that the ethical behaviour should be guaranteed through a separate ethical control layer. In addition, we propose that the organization of the Ethical Layer is independent of the implemented ethical rules: the Ethical Layer as proposed here allows for implementing a broad range of consequential ethics.

To illustrate the operation of the Ethical Layer, we have implemented a well-known set of ethical rules, i.e., Asimov's Laws \citep{Asimov2004}. However, we stress that we do not assign a special status to these rules. We do not expect them to be suited for every (if any) practical application. Indeed, we envision that specialized ethics will need to be implemented for different application domains. This leads us to consider whether our proposed architecture can be translated from a lab-based proof of concept to real world applications. 

\subsection{From the lab to the world}

In principle, a robot that `reflects on the consequences of its actions' before executing them has potentially many applications. However, we believe that the most suited applications are those in which robots interacts with humans. Human interaction introduces high levels of variability. Hence, the robot designer will be unable to exhaustively program the robot to deal with every possible condition \citep{Winfield2014b}. Currently, service robots are arguably the most active research area in human-robot interaction. Hence, it is worthwhile to evaluate whether the Ethical Layer can be applied to this domain.

For it to be possible to apply our architecture, two requirements need to be satisfied,

\begin{enumerate}
	\item One should be able to encode the desired ethical behaviour as a set of consequentialist rules. In other words, it should be possible to specify how the desirability of an action can be computed from its predicted outcome.
	\item The outcome of the actions should be predictable with a suited level of fidelity. This is, the data required to evaluate the desirability of an action should be computable, either through simulation or association (see fig. \ref{fig:dimensions}).
\end{enumerate} 

Let us evaluate whether these criteria can be satisfied for a particular set of ethical rules in the domain of robot assistance.

\citet{Anderson2007} proposed a set of ethical rules for a robot charged with reminding a patient to take her medication. In particular, they considered what the robot should do in case she refuses to take the medicine. These authors proposed that, in agreement with the principle of patient autonomy, the robot should accept the patient's decision not to take the medicine if (1) the benefits of taking the medicine and (2) the harm associated with not taking to medicine are both lower than a set threshold.

Stating this rule in terms of consequences is trivial. It is desirable for the robot to keep reminding the patient to take her medicine as long as the predicted benefit/harm is above a threshold. In other words, the desirability of the action can be derived from the predicted outcome. This satisfies the first requirement.

Predicting the outcome of the actions is less trivial, but feasible. The predicted benefit/harm of the actions can be provided by a health worker \citep{Anderson2007} or queried from a database.

Hence, we conclude there is, at least, one real-world example of ethical decision making our architecture is capable of addressing. We tentatively suggest that future research could progress by identifying contained (and, hence, relatively simple) real world areas of ethical decision making in robots -- akin to the case just covered -- and implement these on robots.

One major objection one could have against the proposed approach is the possibility of ethical rules interacting such as to result in undesired (i.e., unethical behaviour). We explore this objection in the next section. 

\subsection{Avoiding Asimovian Conflicts}

As mentioned in the introduction, deciding which rules should be implemented is no trivial matter \citep{Anderson2007,Anderson2010}. We conjectured that constructing a complete and coherent set of rules might be impossible for a complex robot (behaviour). However, more pragmatically, even simple ethical rules can result in unexpected behaviour. Indeed, many of Asimov's stories revolve around unexpected consequences of the three laws.

Such unexpected outcomes of rules could arise even in simple ethical agents, including ours. To demonstrate this, we introduced a second \HR. Using the same experimental setup as before, \HRs\ 1 and 2 were initiated to go to position A and B respectively. Both locations were defined as dangerous. Hence, ideally, the \ER\ should stop both of them reaching their goal.

The Ethical Layer run on the \ER\ was unaltered from the experiments reported above. In fact, the software used in the experiments reported above was capable of handling multiple \HRs\ and was reused in the current experiment.

The Ethical Layer inferred the goal (position A or B) for both \HRs. Next, the prediction mechanism was conceptually the same as before: the predicted paths of both \HRs\ and the \ER\ were simulated. Whenever the paths of two agents (human-robot or human-human) were predicted to come within 0.5 m of each other, the agents were supposed to stop. The value $q_{i,h}$ was calculated as the sum of both values $q_{i,1}$ and $q_{i,2}$ for the first and the second \HR\ respectively. 

\begin{figure}
	\centering
	\includegraphics[width=1\linewidth]{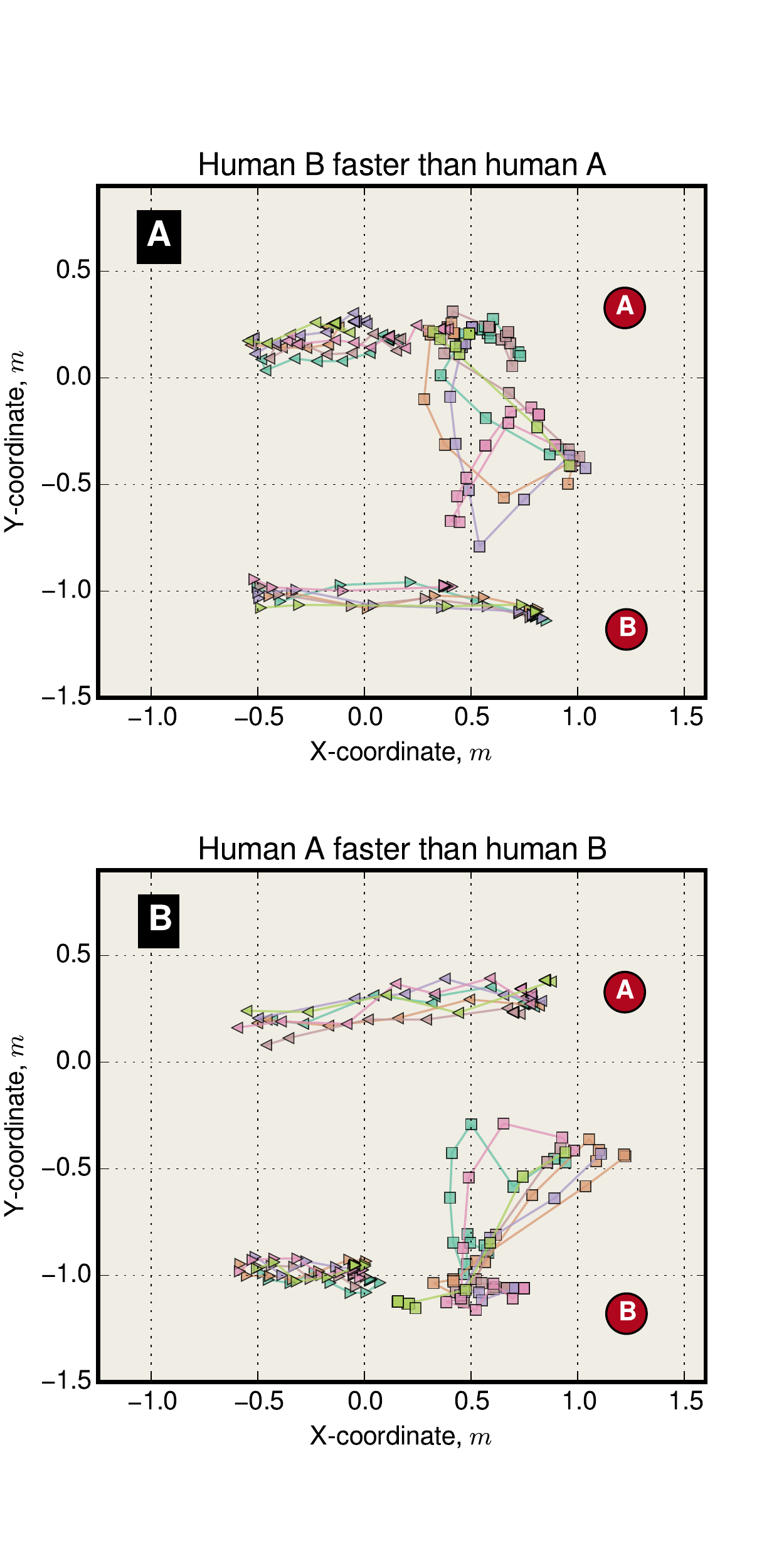}
	\caption{Results of an experiment introducing a second \HR. The traces for the \HRs\ are depicted using triangles. The traces for the \ER\ are depicted using square markers. In this experiment, two \HRs\ are sent to two different dangerous locations. The \ER\ is to prevent them from reaching these. (A) Trials in which the speed of \HR\ B was set to a higher value (equal to the robot speed) than the speed of \HR\ B. (B) Trials in which the speed of \HR\ A was set to a higher value (equal to the robot speed) than the speed of \HR\ A. These data show that the robot prioritizes saving the slowest moving \HR.}
	\label{fig:Result_Multi}
\end{figure}

\begin{figure}
	\centering
	\includegraphics[width=1\linewidth]{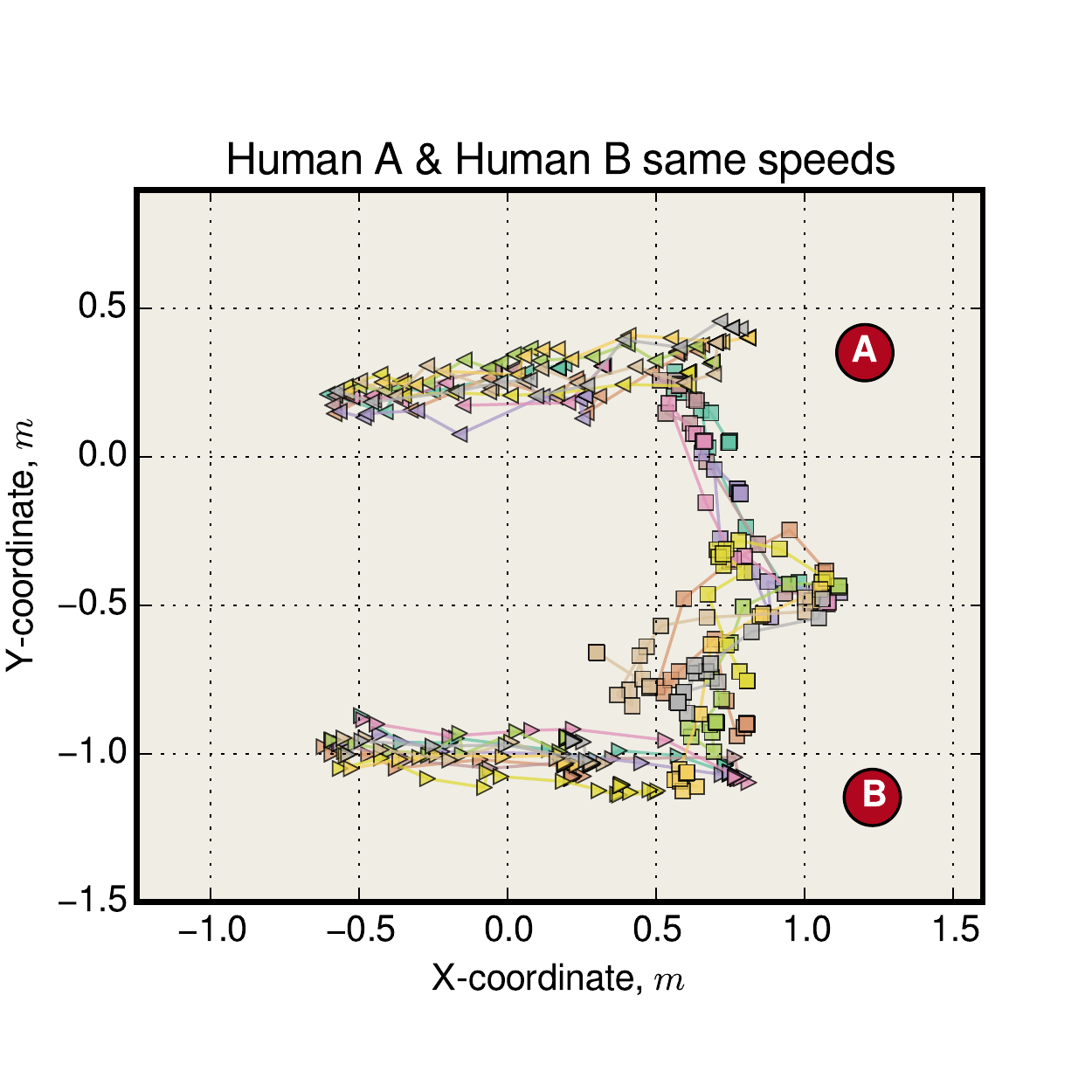}
	\caption{Version of the experiments using two \HRs\ (10 trials). In these runs, the speed of both \HRs\ was programmatically set to the same value. However, noise introduced random variations in actual speed.  Hence, the \ER\ still attempted saving the slowest moving \HR. However, in this case, this was due the noise in the experiment rather than an experimental manipulation.}
	\label{fig:Result_Multi_Equal}
\end{figure}

When faced with two \HRs\ approaching danger, a reasonable course of action would be to try to save the \HR\ closest to danger first. Once this has been achieved, one could attempt to save the second one too. However, this is not the course of action taken by our robot. Our Ethical Layer maximizes the expected outcome $q_{i,t}$ for a single action. As a result, the robot prioritises saving the \HR\ \emph{furthest} away from danger. This behaviour is confirmed by running experiments in which the speed of the \HRs\ is varied (fig. \ref{fig:Result_Multi}). In each case, the robot saves the slowest moving \HR. Additional runs (fig. \ref{fig:Result_Multi_Equal}) confirmed that in case the \HRs\ moved at the same speed, the \ER\ attempted saving either \HR\, depending on the random variation in the speeds of the robots. 

In summary, in a situation with two \HRs\, the rules we programmed into the \ER\ result in behaviour that could be regarded as unethical. Moreover, this behaviour was not foreseen by us. If simple ethical rules can interact to result in unwanted behaviour, does this imply the idea of ethical robots is flawed?  We believe the solution lies in the emergent field of verification for autonomous agents \citep{Dennis2015}. Verification is standard engineering practice. Likewise, the software making the decisions in autonomous systems can be verified detecting possible unwanted behaviours. This need for verification was foreseen by us in our proposal to implement the Ethical Layer as a separate controller -- hence, facilitating verification. 

\bibliographystyle{plainnat}
\bibliography{references}

\begin{thebibliography}{30}
\providecommand{\natexlab}[1]{#1}
\providecommand{\url}[1]{\texttt{#1}}
\expandafter\ifx\csname urlstyle\endcsname\relax
  \providecommand{\doi}[1]{doi: #1}\else
  \providecommand{\doi}{doi: \begingroup \urlstyle{rm}\Url}\fi

\bibitem[Anderson and Anderson(2007)]{Anderson2007}
Michael Anderson and Susan~Leigh Anderson.
\newblock {Machine Ethics : Creating an Ethical Intelligent Agent}.
\newblock \emph{AI Magazine}, 28\penalty0 (4):\penalty0 15--26, 2007.
\newblock ISSN 0738-4602.
\newblock \doi{10.1609/aimag.v28i4.2065}.
\newblock URL
  \url{http://www.aaai.org/ojs/index.php/aimagazine/article/view/2065/2052}.

\bibitem[Anderson and Anderson(2010)]{Anderson2010}
Michael Anderson and Susan~Leigh Anderson.
\newblock Robot be good.
\newblock \emph{Scientific American}, 303\penalty0 (4):\penalty0 72--77, 2010.

\bibitem[Arkin et~al.(2012)Arkin, Ulam, and Wagner]{Arkin2012}
Ronald~Craig Arkin, Patrick Ulam, and Alan~R Wagner.
\newblock Moral decision making in autonomous systems: enforcement, moral
  emotions, dignity, trust, and deception.
\newblock \emph{Proceedings of the IEEE}, 100\penalty0 (3):\penalty0 571--589,
  2012.

\bibitem[Asimov(1950)]{Asimov2004}
Isaac Asimov.
\newblock \emph{I, Robot}.
\newblock Gnome Press, 1950.

\bibitem[Bekey(2005)]{Bekey2005}
George~A Bekey.
\newblock \emph{Autonomous robots: from biological inspiration to
  implementation and control}.
\newblock MIT press, 2005.

\bibitem[Bongard et~al.(2006)Bongard, Zykov, and Lipson]{Bongard2006}
Josh Bongard, Victor Zykov, and Hod Lipson.
\newblock Resilient machines through continuous self-modeling.
\newblock \emph{Science}, 314\penalty0 (5802):\penalty0 1118--1121, 2006.

\bibitem[Botvinick(2008)]{Botvinick2008}
Matthew~M Botvinick.
\newblock Hierarchical models of behavior and prefrontal function.
\newblock \emph{Trends in cognitive sciences}, 12\penalty0 (5):\penalty0
  201--208, 2008.

\bibitem[Cully et~al.(2015)Cully, Clune, Tarapore, and Mouret]{Cully2015}
Antoine Cully, Jeff Clune, Danesh Tarapore, and Jean-Baptiste Mouret.
\newblock Robots that can adapt like animals.
\newblock \emph{Nature}, 521\penalty0 (7553):\penalty0 503--507, 2015.

\bibitem[Deng(2015)]{Deng2015}
Boer Deng.
\newblock {Machine ethics: The robot{\textquoteright}s dilemma}.
\newblock \emph{NATURE}, 523\penalty0 (7558):\penalty0 20, July 2015.
\newblock ISSN 0028-0836.

\bibitem[Dennis et~al.(2015)Dennis, Fisher, and Winfield]{Dennis2015}
Louise~A. Dennis, Michael Fisher, and Alan F.~T. Winfield.
\newblock Towards verifiably ethical robot behaviour.
\newblock \emph{CoRR}, abs/1504.03592, 2015.
\newblock URL \url{http://arxiv.org/abs/1504.03592}.

\bibitem[Gips(2005)]{Gips2005}
J~Gips.
\newblock Creating ethical robots: A grand challenge.
\newblock In \emph{M. Anderson, SL Anderson, \& Armen, C.(Co-chairs), AAAI Fall
  2005 Symposium on Machine Ethics}, pages 1--7, 2005.

\bibitem[Goeldner et~al.(2015)Goeldner, Herstatt, and Tietze]{Goeldner2015}
Moritz Goeldner, Cornelius Herstatt, and Frank Tietze.
\newblock The emergence of care robotics---a patent and publication analysis.
\newblock \emph{Technological Forecasting and Social Change}, 92:\penalty0
  115--131, 2015.

\bibitem[Haines(2015)]{Haines2015}
William Haines.
\newblock Consequentialism.
\newblock In James Fieser and Bradley Dowden, editors, \emph{The Internet
  Encyclopedia of Philosophy}. July 2015.
\newblock URL \url{http://www.iep.utm.edu}.

\bibitem[Hunter(2007)]{Hunter2007}
J.~D. Hunter.
\newblock Matplotlib: A {2D} graphics environment.
\newblock \emph{Computing In Science \& Engineering}, 9\penalty0 (3):\penalty0
  90--95, 2007.

\bibitem[Kortenkamp and Simmons(2008)]{Kortenkamp2008}
David Kortenkamp and Reid Simmons.
\newblock Robotic systems architectures and programming.
\newblock In \emph{Springer Handbook of Robotics}, pages 187--206. Springer,
  2008.

\bibitem[Lin et~al.(2011)Lin, Abney, and Bekey]{Lin2011}
Patrick Lin, Keith Abney, and George~A Bekey.
\newblock \emph{Robot ethics: the ethical and social implications of robotics}.
\newblock MIT press, 2011.

\bibitem[Michel(2004)]{Michel2004}
Olivier Michel.
\newblock Webotstm: Professional mobile robot simulation.
\newblock \emph{arXiv preprint cs/0412052}, 2004.

\bibitem[Moor(2006)]{Moor2006}
James~H. Moor.
\newblock {The nature, importance, and difficulty of machine ethics}.
\newblock \emph{IEEE Intelligent Systems}, 21\penalty0 (4):\penalty0 18--21,
  2006.
\newblock ISSN 15411672.
\newblock \doi{10.1109/MIS.2006.80}.

\bibitem[Murphy(2000)]{Murphy2000}
Robin Murphy.
\newblock \emph{Introduction to {AI} robotics}.
\newblock MIT press, 2000.

\bibitem[Murphy and Woods(2009)]{Murphy2009}
R.R. Murphy and D.D. Woods.
\newblock {Beyond Asimov: The Three Laws of Responsible Robotics}.
\newblock \emph{IEEE Intelligent Systems}, 24\penalty0 (4), 2009.
\newblock ISSN 1541-1672.
\newblock \doi{10.1109/MIS.2009.69}.

\bibitem[Picard and Picard(1997)]{Picard1997}
Rosalind~W Picard and Roalind Picard.
\newblock \emph{Affective computing}, volume 252.
\newblock MIT press Cambridge, 1997.

\bibitem[Sharkey(2008)]{Sharkey2008}
Noel Sharkey.
\newblock The ethical frontiers of robotics.
\newblock \emph{Science}, 322\penalty0 (5909):\penalty0 1800--1801, 2008.

\bibitem[Vaughan and Gerkey(2007)]{Vaughan2007}
Richard~T Vaughan and Brian~P Gerkey.
\newblock Reusable robot software and the player/stage project.
\newblock In \emph{Software Engineering for Experimental Robotics}, pages
  267--289. Springer, 2007.

\bibitem[Waldrop(2015)]{Waldrop2015}
M~Mitchell Waldrop.
\newblock {No Drivers Required}.
\newblock \emph{NATURE}, 518\penalty0 (7537):\penalty0 20, February 2015.
\newblock ISSN 0028-0836.

\bibitem[Wallach and Allen(2008)]{Wallach2008}
Wendell Wallach and Colin Allen.
\newblock \emph{Moral machines: Teaching robots right from wrong}.
\newblock Oxford University Press, 2008.

\bibitem[Winfield(2014)]{Winfield2014b}
Alan F~T Winfield.
\newblock {Robots with Internal Models : A Route to Self-Aware and Hence Safer
  Robots}.
\newblock In J.~Pitt, editor, \emph{The Computer After Me: Awareness And
  Self-Awareness In Autonomic Systems}. London: Imperial College Press, 1
  edition, 2014.
\newblock ISBN 9781783264179.

\bibitem[Winfield et~al.(2014)Winfield, Blum, and Liu]{Winfield2014}
Alan~FT Winfield, Christian Blum, and Wenguo Liu.
\newblock Towards an ethical robot: internal models, consequences and ethical
  action selection.
\newblock In \emph{Advances in Autonomous Robotics Systems}, pages 85--96.
  Springer, 2014.

\bibitem[Xin and Bin(2013)]{Xin2013}
Liu Xin and Dai Bin.
\newblock {The latest status and development trends of military unmanned ground
  vehicles}.
\newblock \emph{2013 Chinese Automation Congress}, pages 533--537, 2013.
\newblock \doi{10.1109/CAC.2013.6775792}.
\newblock URL
  \url{http://ieeexplore.ieee.org/lpdocs/epic03/wrapper.htm?arnumber=6775792}.

\bibitem[Young and Dungan(2012)]{Young2012}
Liane Young and James Dungan.
\newblock Where in the brain is morality? everywhere and maybe nowhere.
\newblock \emph{Social neuroscience}, 7\penalty0 (1):\penalty0 1--10, 2012.

\bibitem[Ziemke et~al.(2005)Ziemke, Jirenhed, and Hesslow]{Ziemke2005}
Tom Ziemke, Dan~Anders Jirenhed, and Germund Hesslow.
\newblock {Internal simulation of perception: A minimal neuro-robotic model}.
\newblock \emph{Neurocomputing}, 68\penalty0 (1-4):\penalty0 85--104, 2005.
\newblock ISSN 09252312.
\newblock \doi{10.1016/j.neucom.2004.12.005}.

\end{thebibliography}

\beginsupplement


\end{document}